\documentclass[runningheads]{llncs}
\usepackage{graphicx}
\usepackage{comment}
\usepackage{amsmath,amssymb} 
\usepackage{color}
\usepackage[dvipsnames]{xcolor}

\DeclareMathOperator{\E}{\mathbb{E}}
\usepackage{soul}

\begin{document}
\pagestyle{headings}
\mainmatter
\def\ECCVSubNumber{5601}  

\title{Matching Embeddings for Domain Adaptation} 

\author{Manuel Pérez-Carrasco\inst{1}\inst{2} \and Guillermo Cabrera-Vives\inst{1} \and Pavlos Protopapas\inst{2} \and Nicolás Astorga\inst{23} \and Marouan Belhaj\inst{2}}
\institute{Department of Computer Science, University of Concepción \\ \email{\{maperezc, guillecabrera\}@udec.cl} \and
Institute for Applied Computational Sciences, Harvard University \\ \email{pavlos@seas.harvard.edu} \and Department of Electrical Engineering, University of Chile \\ \email{nicolas.astorga.r@ing.uchile.cl}
}

\maketitle

\begin{abstract}
In this work we address the problem of transferring knowledge obtained from a vast annotated source domain to a low labeled target domain. We propose Adversarial Variational Domain Adaptation (AVDA), a semi-supervised domain adaptation method based on deep variational embedded representations. We use approximate inference and domain adversarial methods to map samples from source and target domains into an aligned class-dependent embedding defined as a Gaussian Mixture Model. AVDA works as a classifier and considers a generative model that helps this classification. We used digits dataset for experimentation. Our results show that on a semi-supervised few-shot scenario our model outperforms previous methods in most of the adaptation tasks, even using a fewer number of labeled samples per class on target domain.


\end{abstract}

\section{Introduction}
Deep neural networks have become the state of the art for many machine learning problems. However, these methods usually imply the need for a large amount of labeled data in order to avoid overfitting and be able to generalize. Furthermore, it is assumed that the data used for training a model and the data we want to predict come from the same distribution and feature space. This is not always the case, as often the labeled data comes from a \emph{source} domain that is different (but similar) to the \emph{target} domain where we want to predict.
This becomes a huge problem in cases when labeling is costly and/or time-consuming, because the difference between domains (usually known as \emph{domain shift}) can reduce the performance of the predictive models on the target when trained only with data from the source.  One way to address this challenge is to use a source domain which contains a vast amount of annotated data and align this distribution with the target domain, for which we have scarce, or no annotations.


Domain adaptation (DA) methods  aim at reducing the domain shift between datasets \cite{pan_2010}, allowing to generalize a model trained on source to  perform similarly on the target domain by finding a common shared space between them. Deep DA uses deep neural networks to achieve this task. Previous works in deep DA  have addressed the problem of domain shift by using statistical measures such as maximum mean discrepancy \cite{long_2017}, \cite{long_2015}, \cite{tzeng_2014}, \cite{yan_2017}, Kullback-Leibler (KL) divergence \cite{zhuang_2015} and correlation alignment \cite{sun_2016}, \cite{peng_2017}, or introducing class-based loss functions such as softlabel \cite{tzeng_2015}, \cite{gebru_2017}, or contrastive semantic alignment \cite{motiian_2017b} in order to diminish the distance between domain distributions. 

Since the appearance of Generative Adversarial Networks \cite{goodfellow_2014} new approaches have been developed with a focus on Domain Adversarial Neural Networks (DANN) \cite{ganin_2015}, \cite{ganin_2014}. The goal of domain adversarial techniques \cite{tzeng_2017} is to learn a common representation for the source and target data by using a domain discriminator and a feature extractor. The feature extractor is trained to map source and target data into a common representation, while the discriminator is trained to classify data from this representation into source or target. These models are trained by using an adversarial objective. This way, it is possible to find domain-invariant feature space that can be used to solve a classification task on both the source and the target.

Despite domain adversarial methods being good at aligning distributions even in an unsupervised domain adaptation (UDA) scenario (i.e. with no labels from the target), they have problems when facing some domain adaptation challenges. First, since most of these methods were made to tackle UDA problems, they usually fail when there is a significant domain shift between source and target \cite{zou_2019}. Second, in general these methods are not able to take full advantage of the semi-supervised scenario in order to produce more accurate models when a few amount of labels are available from the target \cite{saito_2019}. Labels from the target domain can be used by the model to improve the learning of the classes associated to each unlabeled object, helping to disambiguate the boundaries generated by the source and target alignment. This behavior has been studied in different works, which tried to adapt domain-invariant features from different classes independently \cite{saito_2018}, \cite{kang_2019}.

In this work, we propose Adversarial Variational Domain Adaptation (AVDA), a domain adaptation model which works on a semi-supervised scenario, exploiting labels in the target when they are available by using variational deep embedding (VaDE \cite{jiang_2016}) and domain adversarial methods \cite{ganin_2015}. The idea behind AVDA is to create a common shared latent representation that captures domain-invariant features and class information that allow us to perform classification on target domain. We address this challenge by modeling an embedded space composed by a Mixture of Gaussians, in which samples from the same class are mapped into a Gaussian mixture component, independently of the domain membership. Domain-invariant features are captured directly in this latent space using adversarial learning. Our model also considers a generative process and the alignment of source and target distributions in the latent space. This helps the classification of data for both domains.

The performance of AVDA was validated on benchmark digit recognition tasks  using MNIST \cite{lecunn_1998}, USPS \cite{usps_1988}, and SVHN \cite{svhn_2011} datasets. Through the expermentation, we show that out method outperform previous state-of-the-art in most of cases. Also, our approach has a higher speed of adaptation in the sense that by using a small number of labels in the target domain we are able to obtain competitive results among the current state-of-the-art.

\section{Related Work}

Due the capability of deep networks to learn transferable \cite{bengio_2012}, \cite{yosinski_2014}, \cite{oquab_2014} and invariant \cite{goodfellow_2009} representations of the data, the idea of transferring knowledge acquired from a vast labeled source to increase the performance on a target domain has become a wide area of research \cite{Tan_2018}. Domain adaptation methods deal with this challenge by reducing the domain shift between source and target domains \cite{pan_2010}, aligning a common intern representation for them.

Some statistical metrics have been proposed in order to align source and target distributions, such as maximum mean discrepancy (MMD) \cite{long_2017}, \cite{long_2015}, \cite{tzeng_2014}, \cite{yan_2017}, Kullback Leibler (KL) divergence \cite{zhuang_2015} or correlation alignment (CORAL) \cite{sun_2016}, \cite{peng_2017}. Since the appearance of Generative Adversarial Networks \cite{goodfellow_2014} significant work has been developed around Domain Adversarial Neural Networks \cite{ganin_2015}. The idea of domain adversarial methods is to use a domain classifier which discriminates if a sample belongs to the source or target, while a generator learns how to create indistinguishable representations of data in order to fool the domain classifier. By doing this, a domain-invariant representation of the data distribution is produced in a latent space. 

Despite domain adversarial models achieving good results either by matching distributions in a feature representation (i.e. feature-level) \cite{ganin_2014}, \cite{ming_2017}, \cite{long_2018}, \cite{shu_2018}, \cite{russo_2017}, \cite{zhang_2019} or generating target images that look as if they were part of the source dataset (i.e. pixel-level) \cite{isola_2017}, \cite{zhu_2017}, \cite{hoffman_2017}, \cite{hu_2018}, \cite{hosseini_2019}, when they are used in a UDA scenario, they have difficulties dealing with big domain shifts \cite{zou_2019}. Furthermore, when a few number of annotated target samples are included, these models often do not improve performance relative to just train with unlabeled target samples \cite{saito_2019}. In order to deal with few labels, few-shot domain adaptation methods have been created \cite{motiian_2017a}, \cite{motiian_2017b}, which are not meant to work with unlabeled data, often producing overfitted representations and having problems to generalize on the target domain \cite{wang_2018}.

Semi-supervised domain adaptation (SSDA) deal with these challenges using both labeled and unlabeled samples during training \cite{gong_2012}, \cite{gopalan_2011}, \cite{glorot_2011}, \cite{santos_2017}, \cite{belhaj_2018}, \cite{saito_2018}. Usually for SSDA we are interested on finding a space in which labeled and unlabeled target samples belonging to the same class have a similar internal representation \cite{donahue_2013}, \cite{yao_2015}, \cite{zou_2019}, \cite{saito_2019}. 
A promising approach to deal with labeled and unlabeled samples during training are semi-supervised variational autoencoders \cite{kingma_2014}, \cite{rasmus_2015}, \cite{maloe_2016}. These models seek to map the source and target data into a latent space which depends on the class. As the latent space is shared between labeled and unlabeled data and is class-dependent, points from the same class will be closer in the latent space, helping the classification of unlabeled objects. A promising approach for this task is Variational Deep Embeddings (VaDE \cite{jiang_2016}), which were made for unsupervised clustering but can be easily extended for semi-supervised classification. Through this approach it is possible to map samples from different classes into different Gaussian mixture components, learning class information that can be used for classification.

Our proposed AVDA framework uses VaDE an domain adversarial techniques, mapping source and target data from each class to the same Gaussian component in latent space, allowing the model to use both labeled and unlabeled samples to learn discriminative boundaries for the target domain.

\section{Adversarial Variational Domain Adaptation}

In this work we propose Adversarial Variational Domain Adaptation (AVDA), a model based on semi-supervised variational deep embedding and domain adversarial methods. We use a Gaussian mixture model as a prior for the embedded space and align samples from source and target domains that belong to the same class into the same Gaussian component. 

\subsection{Problem Definition}
In a semi-supervised domain adaptation scenario, we are given a source domain $\mathcal{D}^s = \{\mathbf{x}_{i}^s, y_{i}^s\}_{i=1}^{n^{s}}$ with $n^{s}$ number of labeled samples and a target domain $\mathcal{D}^{u} = \{(\mathbf{x}_{i}^{u})\}^{n^u}_{i=1}$ with $n^{u}$ unlabeled samples. Also, for the target domain we have a subset $\mathcal{D}^{t} = \{(\mathbf{x}_{i}^{t}, y_{i}^{t})\}_{i=1}^{n^t}$ of $n^{t}$ labeled samples. For both domains we have the same $K$ classes, i.e. $y^s_{i}\in\{1,...,K\}$, $y^t_{i}\in\{1,...,K\}$.  Source and target data are drawn from unknown joint distributions $p^{s}(\mathbf{x}^{s},y^{s})$ and $p^{t}(\mathbf{x}^{t},y^{t})$ respectively, where $p^{s}(\mathbf{x}^{s},y^{s}) \neq p^{t}(\mathbf{x}^{t},y^{t})$. 

The goal of this work is to build a model that provides an embedding space $\mathbf{z}$ in which source and target data have the same representation for each of the $K$ classes, in cases where $n^t$ is small (i.e few-shot scenario). We propose the use of a Semi-Supervised Variational Deep Embedding to model the latent space distribution \cite{jiang_2016}.

\subsection{Adversarial Variational Domain Adaptation Model}

\begin{figure*}[t]
  \includegraphics[width=\linewidth]{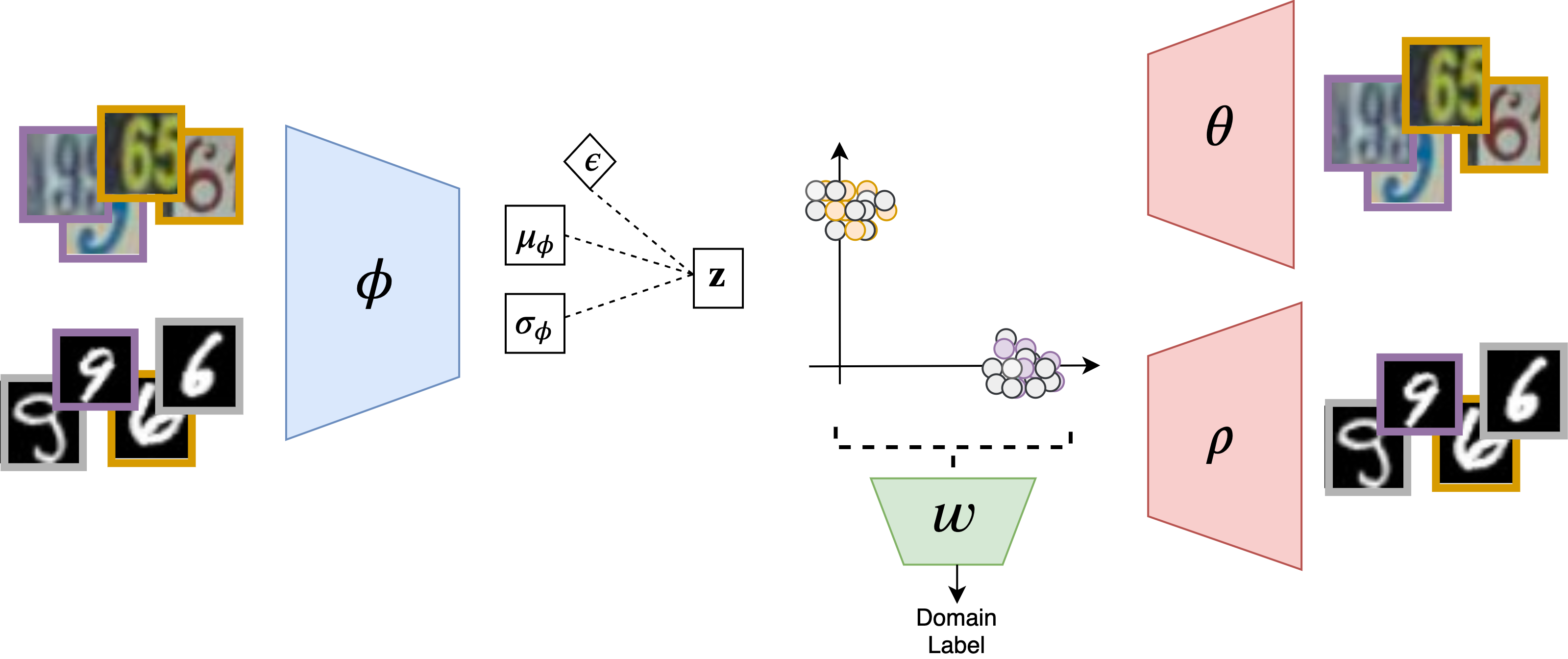}
  \caption{Overall architecture for Adversarial Variational Domain Adaptation (AVDA). The model is composed of an encoder defined by the parameter $\phi$, two decoders defined by parameters $\theta$ and $\rho$, for source and target respectively, and a discriminator on the latent space described by parameters $w$. The source and target embedded spaces are modeled as mixture models where each class is mapped to the same component. The goal of AVDA is to generate an aligned latent space for accurately classify both source and target data by minimizing the difference between their latent representations. 
  }
  \label{AVDA}
   \vspace{-1.0em}

\end{figure*}

We use a single encoder for both the source and target domain defined by parameters $\phi$. 
This  neural network will encode the data into a source embedded space $\mathbf{z}^s$ and a target embedded space $\mathbf{z}^t$. We model the distribution of the latent space $\mathbf{z}^s$ and $\mathbf{z}^t$ as Gaussian Mixture Models where objects of each class are mapped to a different Gaussian component. We sample from the latent spaces and use a source and a target decoder parameterized by $\theta$ and $\rho$ respectively. Our goal is to match the source and target latent space so that data from the same class are mapped to  matched Gaussian component. Furthermore, we expect this latent space to be separable so we can predict class labels for both the source and target domains by calculating the probability of the sample of belonging to each Gaussian component given the class.  For extracting domain-invariant features, we regularize our latent variables via adversarial training (i.e optimizing a domain discriminator using the latent variables and optimizing the encoder to make this discriminator unable to distinguish if the latent variable comes from the source or target distribution). The overall model is displayed in Figure \ref{AVDA}.



The generative processes for the source and target are defined by the joint probability distributions: 
\begin{align}
    p(\mathbf{x}^s, y^s, \mathbf{z}^s) &= p(y^s)p(\mathbf{z}^s|y^s)p(\mathbf{x}^s|\mathbf{z}^s),\\
    p(\mathbf{x}^t, y^t, \mathbf{z}^t) &= p(y^t)p(\mathbf{z}^t|y^t)p(\mathbf{x}^t|\mathbf{z}^t),\end{align}
where  we define each probability as follows:

\begin{align}
p(y^{s}) &= Cat(y^{s}|\pi^s), \\
p(\mathbf{z}^{s}|y^{s}) &= \mathcal{N}(\mathbf{z}^{s}|\mu(y^s), \sigma^{2}(y^s)\mathbf{I}),  \\
p_{\theta}(\mathbf{x}^{s}|\mathbf{z}^{s}) &= Ber(\mathbf{x}^{s}|\mu_{x}(\mathbf{z}^{s},\theta))\,\ or \,\ \mathcal{N}(\mathbf{x}^{s}|\mu_{x}(\mathbf{z}^{s},\theta), \sigma_{x}^{2}(\mathbf{z}^{s},\theta)\mathbf{I}),
\end{align}
\begin{align}
p(y^{t}) &= Cat(y^{t}|\pi^t), \\ 
p(\mathbf{z}^{t}|y^{t}) &= \mathcal{N}(\mathbf{z}^{t}|\mu(y^s), \sigma^{2}(y^s)\mathbf{I}), \\
p_{\rho}(\mathbf{x}^{t}|\mathbf{z}^{t}) &= Ber(\mathbf{x}^{t}|\mu_{x}(\mathbf{z}^{t},\rho))\,\ or \,\ \mathcal{N}(\mathbf{x}^{t}|\mu_{x}(\mathbf{z}^{t},\rho), \sigma_{x}^{2}(\mathbf{z}^{t},\rho)\mathbf{I}),
\end{align}
where $Cat(y|\pi)$ is a categorical distribution parametrized by $\pi$, the prior probability for class $y$, $\pi \in \mathcal{R}_{+}^{K}$, $\sum_{y=1}^{K}\pi_{i}=1$. At the same time, $\mu(y)$ and $\sigma^{2}(y)$ are the mean and variance of the embedded normal distribution corresponding to class labels $y$. $\mathbf{I}$ is the identity matrix. The likelihood distributions for the generative process are defined as a Bernoulli for a binary output or a Gaussian distribution for a continuous output, whose parameters are obtained using neural networks parametrized by $\theta$ and $\rho$ for the source and target domains respectively.
On a semi-supervised scenario we optimize an objective composed of functions for the labeled and unlabeled data (see Section \ref{sec:variation_objectives}). 
We aim at approximate the marginal log-likelihood of the data distribution by maximizing a variational lower bound that depends on an approximated posterior distribution $q_{\phi}(y,\mathbf{z}|\mathbf{x})$ (recall the encoder with parameters $\phi$ is used for both source and target data, whereby the superscript are omitted).
We assume that this distribution can be factorized as $q(y,\mathbf{z}|\mathbf{x}) = q(\mathbf{z}|\mathbf{x})q(y|\mathbf{x})$ ($y$ and $\mathbf{z}$ are conditionally independent given $\mathbf{x}$), and model it by using normal and categorical distributions as follows:
\begin{align}\label{eq:q_phi}
q_{\phi}(\mathbf{z}|\mathbf{x}) &= \mathcal{N}(\mathbf{z}|\mu_{\phi}(\mathbf{x}), \sigma_{\phi}^{2}(\mathbf{x})\mathbf{I}), \\
q_{\phi}(y|\mathbf{x}) &= Cat(y|\pi_{\phi}(\mathbf{x})), \end{align}
where $\mu_{\phi}(\mathbf{x})$ and $\sigma_{\phi}^{2}(\mathbf{x})$ are the outputs of a neural network parametrized by $\phi$. It is possible to sample $\mathbf{z}$ and perform backpropagation by using the reparametrization trick as defined in \cite{kingma_2013}. We assume that by modeling $p(\mathbf{z}|y)$ as a Gaussian ditribution, we will be able to learn discriminative features, being able to obtain the parameters $\pi(\mathbf{z})$ that model a categorical distribution for the classes $y \in {1, ..., K}$.  


\subsection{Variational Objective}\label{sec:variation_objectives}

Deep neural network models have been used to learn continuous latent variable models with intractable posterior distributions using variational inference \cite{kingma_2013}, \cite{rezende_2014}. The idea of variational inference is to approximate the true posterior distribution $p(y,\mathbf{z}|\mathbf{x})$ by  using a simpler approximated posterior distribution  $q(y,\mathbf{z}|\mathbf{x})$. Through this approximation we will be able to map samples into a known distribution in latent space, and sample from this distribution in order to generate new data using the generative model. In this work, we take advantage of this known distribution in order to map samples from the same class into the same distribution. This is done through the using of the variational objectives.

\subsubsection{Supervised Source Objective}

For the source domain, we aim at learning a latent space that captures class information, modeled by using a Gaussian mixture model. We then use that embedded space to generate data that looks like our training set through a decoder parametrized by $\theta$. This is achieved by maximizing the lower bound of the marginal log-likelihood for the labeled data as follows:
\begin{equation}
\log{p(\mathbf{x},y)} \begin{aligned}[t] & = \log \int_{\mathbf{z}} p(\mathbf{x},y,\mathbf{z})d\mathbf{z} \\
 & \geq  \mathbb{E}_{q_{\phi}(\mathbf{z}|\mathbf{x})}\Big[\log {\frac{p(\mathbf{x},y,\mathbf{z})}{q_{\phi}(\mathbf{z}|\mathbf{x})}\Big]} \\
& =  \mathbb{E}_{q_{\phi}(\mathbf{z}|\mathbf{x})}[\log p_{\theta}(\mathbf{x}|\mathbf{z})] - D_{KL}(q_{\phi}\big(\mathbf{z}|\mathbf{x})||p(\mathbf{z}|y)\big) + \log{p(y)} \\
 & = \mathcal{L}^{s}_{ELBO},
\end{aligned}
 \label{eq:Lelbo_s}
\end{equation}
Since $\log{p(y)}$ is a constant,
minimizing this lower bound is the same as minimizing the Kullback-Leibler Divergence between $q_{\phi}(\mathbf{z}^{s}|\mathbf{x}^{s})$ and the class-dependent prior distribution that we defined in Equation 4, and computing the maximum likelihood estimation of $\mathbb{E}_{q_{\phi}(\mathbf{z}|\mathbf{x})}[\log p_{\theta}(\mathbf{x}|\mathbf{z})]$, which measures the reconstruction quality. Then, the supervised objective for the source domain can be optimized by minimizing the following objective:
\begin{equation}
 \mathcal{L}^{s}_{sup} = - \mathcal{L}^{s}_{ELBO}  + \alpha^{s}\mathbb{E}_{(\mathbf{x},y)\sim \mathcal{D}^{s}}[-\log{q_{\phi}(y|\mathbf{x})}],
 \label{eq:Lsup_s}
\end{equation}
where a discriminative term $\mathbb{E}_{(\mathbf{x},y)\sim \mathcal{D}^{s}}[\log{q_{\phi}(y|\mathbf{x})}]$ was included in order to generate discriminative boundaries around the Gaussian mixture components. $\alpha^{s}$ is the hyper-parameter that controls the relative importance of the discriminative process in the model.


\subsubsection{Supervised Target Objective}

Our main objective is to find a latent space in which source and target follow the same distribution. For that purpose we use the same encoder for both, which is parametrized by $\phi$ which we expect will capture domain-invariant features along with class information. At the same time, we use a decoder parametrized by $\rho$ to generate samples that look like the target data. Following a similar procedure as with the source domain, we can derive a lower bound for the marginal log-likelihood for target datapoints as follows:

\begin{equation}
\log{p(\mathbf{x},y)} \begin{aligned}[t] & = \log \int_{\mathbf{z}} p(\mathbf{x},y,\mathbf{z})d\mathbf{z} \\
& \geq  \mathbb{E}_{q_{\phi}(\mathbf{z}|\mathbf{x})}\Big[\log {\frac{p(\mathbf{x},y,\mathbf{z})}{q_{\phi}(\mathbf{z}|\mathbf{x})}\Big]} \\
& =  \mathbb{E}_{q_{\phi}(\mathbf{z}|\mathbf{x})}[\log p_{\rho}(\mathbf{x}|\mathbf{z})] - D_{KL}(q_{\phi}\big(\mathbf{z}|\mathbf{x})||p(\mathbf{z}|y)\big) + \log{p(y)} \\
 & = \mathcal{L}^{t}_{ELBO}.
\end{aligned}
 \label{eq:Lelbo_t}
\end{equation}
which is similar to Equation \ref{eq:Lsup_s}, but differs in the likelihood $\mathbb{E}_{q_{\phi}(\mathbf{z}|\mathbf{x})}[\log p_{\rho}(\mathbf{x}|\mathbf{z})]$, which measures the reconstruction quality for the target datapoints using the decoder parametrized by $\rho$. The supervised objective for the labeled target datapoints can be written as:
\begin{equation}
 \mathcal{L}^{t}_{sup} = - \mathcal{L}^{t}_{ELBO}  + \alpha^{t}\mathbb{E}_{(\mathbf{x},y)\sim \mathcal{D}^{t}}[-\log{q_{\phi}(y|\mathbf{x})}],
 \label{eq:Lsup_t}
\end{equation}
where we also include a discriminative term $\mathbb{E}_{(\mathbf{x},y)\sim \mathcal{D}^{t}}[\log{q_{\phi}(y|\mathbf{x})}]$, thus encouraging discriminative boundaries in the latent space defined by the target samples. $\alpha^{t}$ is the hyper-parameter that controls the relative importance of the discriminative process in the model.

\subsubsection{Unsupervised Objective}\label{sec:u_target_obj}
The unsupervised variational lower bound for the unlabeled target datapoints can be derived by noticing that since we do not have labels, we can treat $z$ and $y$ as latent variables as follows:

\begin{equation} \label{eq:Lunsup_elbo}
  \log{p(\mathbf{x})}  \begin{aligned}[t]  &= \log{\int_{z}\sum_{k} p(\mathbf{x},y,\mathbf{z})d\mathbf{z}} \\
  &\geq  \mathbb{E}_{q_{\phi}(y,\mathbf{z}|\mathbf{x})}\Big[\log {\frac{p(\mathbf{x},y,\mathbf{z})}{q_{\phi}(\mathbf{z},y|\mathbf{x})}\Big]} \\
  &= \mathbb{E}_{q_{\phi}(y,\mathbf{z}|\mathbf{x})}[\log p_{\rho}(\mathbf{x}|\mathbf{z})] - D_{KL}(q_{\phi}\big(y,\mathbf{z}|\mathbf{x})||p(\mathbf{z},y)\big) \\
  &= \mathbb{E}_{q_{\phi}(y,\mathbf{z}|\mathbf{x})}\Big[\log p_{\rho}(\mathbf{x}|\mathbf{z}) - \log \frac{q_{\phi}(\mathbf{z}|\mathbf{x})}{p(\mathbf{z}|y)} -  \log \frac{q_{\phi}(y|\mathbf{x})}{p(y)} \Big] \\
  & = \mathcal{L}^{u}_{ELBO} \\ &= - \mathcal{L}_{unsup}.
\end{aligned}
\end{equation}
Intuitively, the term $\mathbb{E}_{q_{\phi}(\mathbf{z}|\mathbf{x})}[\log p_{\rho}(\mathbf{x}|\mathbf{z})]$ aims at achieving a better reconstruction, and $D_{KL}(q_{\phi}\big(\mathbf{z},y|\mathbf{x})||p(\mathbf{z},y)\big)$ regularizes the posterior distribution $q_{\phi}\big(\mathbf{z},y|\mathbf{x})$ by following the prior $p(\mathbf{z},y)$.

Following Jiang et al. 2017 \cite{jiang_2016}, the variational lower bound can be rewritten as:
\begin{equation}
     \mathcal{L}^{u}_{ELBO} = \int_{\mathbf{z}} q(\mathbf{z}|\mathbf{x})\log{\frac{p(\mathbf{x}|\mathbf{z})p(\mathbf{z})}{q(\mathbf{z}|\mathbf{x})}}d\mathbf{z} - \int_{\mathbf{z}} q(\mathbf{z}|\mathbf{x}) D_{KL}\big(q(y|\mathbf{x})||p(y|\mathbf{z})|\big)d\mathbf{z}.
\end{equation}\label{eq:Lu}
The first term on Eq. \ref{eq:Lu} does not depend on $y$, and as the Kullback-Leibler divergence in non negative, for this lower bound to be maximized, it must be given that $D_{KL}\big(q(y|\mathbf{x})||p(y|\mathbf{z})|\big) = 0$, which implies that $q(y|\mathbf{x}) = p(y|\mathbf{z})$. As a consequence of this restriction, we can compute $q(y|x)$ as:
\begin{equation} \label{discriminative}
   q(y|\mathbf{x}) = p(y|\mathbf{z}) = \frac{p(\mathbf{z}|y)p(y)}{\sum_{k=1}^{K}p(\mathbf{z}|y=k)p(y=k)} .
\end{equation}
From this equation we can obtain the class for the unlabeled datapoints directly from the Gaussian mixture components. As our objective is to create an aligned latent space from which we can classify new unseen target datapoints, we optimize the discriminative terms introduced in Equation \ref{eq:Lsup_s} and Equation \ref{eq:Lsup_t} by using Equation \ref{discriminative}.

\subsubsection{Optimization}
The overall variational objective can be optimized as follows:
\begin{equation}
\min_{\phi, \theta, \rho} \mathcal{L}^{v} = \gamma \mathcal{L}_{sup} + (1-\gamma) \mathcal{L}_{unsup},
\label{eq:Lvar}
\end{equation}
where $\mathcal{L}_{sup} = \mathcal{L}^{s}_{sup} + \mathcal{L}^{t}_{sup}$, and $\gamma$ is a hyperparameter that controls the relative importance between labeled and unlabeled datapoints. Assuming a Gaussian distribution for $\mathbf{x}$, it is possible to rewrite $\mathcal{L}_{sup}$ as follows (see Appendix A for details):

\begin{equation} \label{eq:lsup_1}
\mathcal{L}_{sup} \begin{aligned}[t] =  & \sum^{D}_{i=1}
\log{\sigma_{x}|_{i}} + \frac{(x_{i} - \mu_{x}|_{i})^{2}}{2\sigma^{2}|_{i}} \\
& + \frac{1}{2}\sum^{J}_{j=1} \Big(\log{(\sigma^{2}(y)|_{j})} + \frac{\sigma^{2}_{\phi}(\mathbf{x})|_{j}}{\sigma^{2}(y)|_{j}} + \frac{(\mu_{\phi}(\mathbf{x})|_{j}-\mu(y)|_{j})^{2}}{\sigma^{2}(y)|_{j}} \Big) \\
& - \sum^{J}_{j=1}(1+\log \sigma^{2}(\mathbf{x})|_{j}) + H(q_{\phi}(y|\mathbf{x}), y),
\end{aligned}
\end{equation}
where $D$ is the dimensionality of $\mathbf{x}$ and $\mathbf{\mu}_{x}$, $x_{i}$ is the $i^{th}$ element of $\mathbf{x}$, $J$ is the dimensionality of the latent space and $*|_{j}$ denotes the $j^{th}$ element of $*$. $H(\cdot,\cdot)$ is the categorical cross entropy loss, $q_{\phi}(y|\mathbf{x})$ can be obtained from Equation \ref{discriminative}, and $y$ is the known label for the supervised case. In a similar way, the unsupervised objective can be rewritten as (see Appendix B for details):
\begin{equation} \label{eq:lunsup_1}
\mathcal{L}_{unsup} \begin{aligned}[t] =  & \sum^{D}_{i=1} \log{\sigma_{x}|_{i}} + \frac{(x_{i} - \mu_{x}|_{i})^{2}}{2\sigma^{2}|_{i}} \\
&  + \frac{1}{2}\sum^{K}_{k=1}q_{\phi}(y_{k}|\mathbf{x})\sum^{J}_{j=1} \Big(\log{\sigma^{2}(y_{k})|_{j})} + \frac{\sigma^{2}_{\phi}(\mathbf{x})|_{j}}{\sigma^{2}(y_{k})|_{j}} + \frac{(\mu_{\phi}(\mathbf{x})|_{j}-\mu(y_{k})|_{j})^{2}}{\sigma^{2}(y_{k})|_{j}} \Big) \\
& -\sum^{K}_{k=1}q_{\phi}(y_{k}|\mathbf{x}) \log{\frac{\pi_{k}}{q_{\phi}(y_{k}|\mathbf{x})}} - \sum^{J}_{j=1}(1+\log \sigma^{2}(\mathbf{x})|_{j}).
\end{aligned}
\end{equation}
The parameter $\mathbf{\mu}_{x}$ and $\sigma^{2}_{\mathbf{x}}$ for the Gaussian distribution of $\mathbf{x}$ are computed different if the sample comes from source or target domain. For source domain we compute the parameters for the likelihood distribution as:

\begin{equation}
    [\mu^{s}_{\mathbf{x}}, \log{\sigma}^{s}_{\mathbf{x}}] = f(\mathbf{z}^{s}; \theta),
\end{equation}
where $f$ is the neural network that we defined for the decoder, which takes $\mathbf{z}^{s} \sim q_{\phi}(\mathbf{z}^{s}|\mathbf{x}^{s})$ as input and uses the parameters $\theta$ to generate $\mu^{s}_{x}$ and $\log{\sigma}^{s}_{x}$. In a similar way, the parameters  $\mu^{t}_{\mathbf{x}}$ and $\log{\sigma}^{t}_{\mathbf{x}}$ for the target domain are computed as:
\begin{equation}
    [\mu^{t}_{\mathbf{x}}, \log{\sigma}^{t}_{\mathbf{x}}] = f(\mathbf{z}^{t}; \rho),
\end{equation}
where we use a different decoder which takes $\mathbf{z}^{t} \sim q_{\phi}(\mathbf{z}^{t}|\mathbf{x}^{t})$ as input and uses the parameters $\rho$ to generate $\mu^{t}_{\mathbf{x}}$  and $\log{\sigma}^{t}_{\mathbf{x}}$. In both cases, we can optimize this parameter by sampling $\mathbf{z}$ using the reparametrization trick defined in \cite{kingma_2013} and \cite{rezende_2014}, where $\mathbf{z}$ is obtained by:
\begin{equation} \label{reparam}
    \mathbf{z} = \mu_{\phi}(\mathbf{x}) + \sigma^{2}_{\phi}(\mathbf{x}) \odot \epsilon,
\end{equation}
where $\epsilon \sim \mathcal{N}(0,\mathbf{I})$ and $\odot$ is the element-wise multiplication. This reparametrization trick allows us to backpropagate the errors through the continuous latent variable $\mathbf{z}$.


\subsection{Adversarial Objective}
We would like the marginal
distribution over the approximated posterior distribution $q(\mathbf{z})=\E_{p(\mathbf{x})}{[q(\mathbf{z}|\mathbf{x})]}$ to be the same for the source and the target domains (i.e $q(\mathbf{z}^{s}) = q(\mathbf{z}^{t}) \equiv q(\mathbf{z})$). For this purpose, we use generative adversarial learning \cite{goodfellow_2014} in order to encourage the alignment between distributions and force the encoder to capture domain-invariant features \cite{ganin_2015}. In particular, we use a discriminator $D_{w}(\cdot)$ with parameters $w$ which takes the form of a classifier distinguishing between source and target data. We used a domain label which was defined as $1$ for source data and $0$ for target data. By doing this, we encourage the discriminator to learn the underlying distribution for the latent variables of each domain. This discriminator, can be optimized with respect to the parameters $w$ as follows:

\begin{equation} \label{eq:LD}
   \min_{w} \mathcal{L}_{D} = -\E_{ q_{\phi}(\mathbf{z}^{s}|\mathbf{x}^{s})}[\log D_{w}(\mathbf{z}^{s})] - \E_{q_{\phi}(\mathbf{z}^{t}|\mathbf{x}^{t})}[\log (1- D_{w}(\mathbf{z}^{t}))],
\end{equation}
where z are sampled following the reparametrization trick described in Equation \ref{reparam}.
On the the other hand, we want the encoder to be able to create indistinguishable latent representations for source and target domains by capturing domain-invariant features. We can achieve this task by training and adversarial objective with respect to the encoder parameters $\phi$ as follows:
\begin{equation} \label{eq:LD}
    \min_{\phi} \mathcal{L}_{A} = \E_{q_{\phi}(\mathbf{z}^{t}|\mathbf{x}^{t})}[\log (1 - D_{w}(\mathbf{z}^{t}))].
\end{equation}
In practice, AVDA is trained through alternate optimization of the described Variational and Adversarial objectives.

\section{Experiments}

We evaluate our framework on the digits dataset composed of MNIST \cite{lecunn_1998}, USPS \cite{usps_1988}, and Street View House Numbers (SVHN) \cite{svhn_2011}. We perform different adaptations between those datasets varying the number of labeled samples used on the target domain, as well as using different number of labeled source samples.


\subsection{Dataset}

 We use three benchmark digits datasets: MNIST (\textbf{M}), USPS (\textbf{U}), and Street View House Numbers (SVHN) (\textbf{S}). Theses datasets contain images of digits from 0 to 9 in different environments: \textbf{M} and \textbf{U} contain handwritten digits while, \textbf{S} contain natural scene images. Each dataset is composed of its own train and test set: 60,000 train samples and 10,000 test samples for MNIST;  7,291 train samples and 2,007 test samples for USPS; and 73,257 train samples and 26,032 test samples samples for SVHN. 


\subsection{Implementation Details}
For all the experiments, images were rescaled to $[-1.0, 1.0]$ and resized to $32$x$32$. The SVHN dataset contains RGB images, so for SVHN-MNIST adaptation, the MNIST images were repeated in each of the three filters in order to use the same input tensor size to the network. We follow \cite{saito_2019} and select three labeled examples from the target domain as a validation set. The hyperparameters were empirically selected by measuring the accuracy of our model on this validation set. For all scenarios we set the parameter that control the importance of the discriminative model as $\alpha^{s}=1$ and $\alpha^{t}=1$. $\gamma$ was set as $0.9$. Our embedding is created on a 20-dimensional space, where the means $\mu_\phi$ and standard deviations $\sigma_\phi$ of each Gaussian component are learned via backpropagation. Similarly to Jiang et al. 2017 \cite{jiang_2016}, we pretrain AVDA by iteratively using the supervised variational objective for the source domain and only the reconstruction objective for the target $\E_{q_\phi(y, \mathbf{z}|\mathbf{x}))}[\log p_\rho(\mathbf{x}|\mathbf{z})]$ (1 batch at the time).  This pretraining allows us to initialize the mean and variance for the prior distributions and also avoid to getting stuck in an undesirable local minima due to the reconstruction term.
Our training was performed using the Adam optimizer \cite{kingma_2015} with parameters $\beta_{1} =0.9$ and $\beta_{2} = 0.999$ and a learning rate of $0.0001$. We used mini-batches of 128 samples for source, labeled target and unlabeled target samples. If less than 128 samples are available in the target labeled data, we use a batch of the total number of such samples for the supervised target objective. We do a forward pass using these three mini-batches (source, labeled target and unlabeled target), calculate their respective objectives, add the losses and do a backward pass. We call each of these forward-backward passes an iteration. We decreases the learning rate by a decay rate of $0.9$ every 100 iterations. 

\subsection{Results}
We compare our results against the following state-of-the-art methods for supervised and semi-supervised domain adaptation: CCSA \cite{motiian_2017b}, FADA \cite{motiian_2017a}, F-CADA \cite{zou_2019} and $d$-SNE \cite{Xu_2019_CVPR}.
For the first experiment we used the evaluation protocol  defined in CCSA \cite{motiian_2017b}: 2.000 samples where randomly selected from MNIST as a source, and 0 labels (denoted as 0-shot),  one label (denoted as 1-shot), three labels per class (denoted as 3-shot), five labels per class (denoted as 5-shot) and seven labels per class (denoted as 7-shot) where chosen from USPS as target. For fair comparison we used LeNet \cite{lecunn_1998} as the network architecture for the encoder. The results are shown in Table \ref{digits_table_semisup_2000}. As can be seen, our proposed method outperform previous state of the art in all the cases for this task even in the unsupervised scenario (i.e. 0 labels per class).

For the second experiment we used the evaluation protocol proposed in F-CADA \cite{zou_2019}, i.e. three domain adaptation task are performed: from MNIST to USPS (\textbf{M$\rightarrow$ U}), from USPS to MNIST (\textbf{U$\rightarrow$ M}), and from SVHN to MNIST (\textbf{S$\rightarrow$ M}). We performed 5 experiments using 1-shot, 3-shot, 5-shots, 7-shot and 10-shot. The rest of the training samples are used in an unsupervised fashion (Notice that CCSA \cite{motiian_2017b} and FADA \cite{motiian_2017a} do not utilize unlabeled samples during training while we do). We use similar network architectures as the ones proposed in \cite{ming_2017}, \cite{hu_2018}. The results are displayed in Table \ref{digits_table_semisup}. Notice we report two results for our method, named \emph{best} and \emph{random}. As we use stochastic optimization, results will depend on the randomness of the training procedure. We trained our model 5 times and report as \emph{random} the mean and standard deviation of the accuracy over the test set. The value reported as \emph{best} is the accuracy over the test set of the best model in terms of the evaluation over the validation set. For the 1-shot and 3-shot tests we outperform all previous approaches. For the 5-shot test F-CADA performs better than AVDA only on the \textbf{U$\rightarrow$ M} task. For the 7-shot and 10-shot test, our model still performs better than the rest on the \textbf{S$\rightarrow$ M} task, and achieve competitive results for the \textbf{M$\rightarrow$ U} and \textbf{U$\rightarrow$ M} tasks. In general, our approach has a higher speed of adaptation in the sense that by using a small number of labels in the target domain we are able to obtain competitive results.

\begin{table}
  \begin{center}
  \begin{tabular}{|l|c|c|c|c|c|}
    \hline

     Method  &   0-shot & 1-shot     & 3-shot & 5-shot & 7-shot  \\ \hline
    \hline
    CCSA \cite{motiian_2017b} & 65.40 & 85.00 & 90.10 &92.40 & 92.90 \\
    FADA \cite{motiian_2017a} & 65.40 & 89.10 & 91.90 & 93.40 & 94.40  \\
    $d$-SNE \cite{Xu_2019_CVPR} & 73.01  & 92.90 & 93.55 & 95.13 & 96.13  \\

\hline
    AVDA (ours)  & $\boldsymbol{97.34}$ & $\boldsymbol{97.54}$ & $\boldsymbol{97.71}$ & $\boldsymbol{97.80}$ & $\boldsymbol{97.83}$ \\ 
    \hline    
  \end{tabular}
    \caption{Results on \textit{Digits} dataset for semi-supervised task using 2000 datapoints in the source domain.}
  \label{digits_table_semisup_2000}
  \vspace{-1.0em}
  \end{center}
\end{table}

\begin{table}[h]
  \begin{center}
  \scalebox{0.75}{
  \begin{tabular}{|l|ccc|ccc|ccc|ccc|ccc|}
    \hline
    &\multicolumn{3}{c|}{1-shot}&
    \multicolumn{3}{c|}{3-shot}&
    \multicolumn{3}{c|}{5-shot}&
    \multicolumn{3}{c|}{7-shot}&
    \multicolumn{3}{c|}{10-shot}   \\

    Method     & \textbf{A}     & \textbf{B} & \textbf{C} & \textbf{A}     & \textbf{B} & \textbf{C} & \textbf{A}     & \textbf{B} & \textbf{C} & \textbf{A}     & \textbf{B} & \textbf{C} & \textbf{A}     & \textbf{B} & \textbf{C} \\ \hline

    CCSA \cite{motiian_2017b}  & 85.00 & 78.40 & - & 90.10 & 85.80 & - & 92.40 & 88.80 & - & - & - & - & 97.27 & 95.71 & 92.40\\
    FADA \cite{motiian_2017a}  & 89.10 & 81.10 & 72.80 & 91.90 & 87.50 & 82.60 & 93.40 & 91.10 & 86.10 & 94.40 & 91.50 & 87.20 & - & - & -\\
    F-CADA \cite{zou_2019}  & 97.20 & 97.50 & 94.80 & 97.90 & 98.10 & 95.40 & 98.30 & $\boldsymbol{98.60}$ & 95.60& $\boldsymbol{98.60}$ & $\boldsymbol{98.90}$ & 96.61 & - & - & -
\\
    $d$-SNE \cite{Xu_2019_CVPR}  & - & - & - & - & - & - & - & - & - & 97.53 & 97.52 & 95.68 & $\boldsymbol{99.00}$ & 98.49 & 96.45 \\

\hline
    AVDA (ours)  & $\boldsymbol{98.23}$       & $\boldsymbol{98.38}$ & $\boldsymbol{96.60}$ 
    & $\boldsymbol{98.26}$ & $\boldsymbol{98.39}$ & $\boldsymbol{97.28}$
                   & $\boldsymbol{98.43}$       & 98.43 & $\boldsymbol{97.56}$ & 98.51 & 98.43 & $\boldsymbol{97.69}$ & 98.39 & $\boldsymbol{98.57}$ & $\boldsymbol{97.67}$\\ 
best & & & & & & & & & & & & & & &\\
    \hline        
    AVDA (ours) & 98.05      & 97.5 & 96.09 
    & 98.11 & 98.36 & 97.19
                   & 98.25       & 98.37 & 97.41 & 98.38 & 98.39 & 97.43 & 98.31 & 98.48 & 97.46\\
                random   & $\pm0.13$ & $\pm0.7$ & $\pm0.62$ & $\pm0.07$ &$\pm0.02$ & $\pm0.06$
            
                   & $\pm0.10$       & $\pm0.07$ & $\pm0.09$ & $\pm0.06$ & $\pm0.04$ & $\pm0.07$ & $\pm0.10$  & $\pm0.06$ & $\pm0.06$\\

    \hline    
  \end{tabular}}
    \caption{Results on \textit{Digits} dataset for semi-supervised task. All the results are reported using accuracy by performing 5 random experiments. For this table \textbf{M$\rightarrow$ U} is denoted as \textbf{A}, \textbf{U$\rightarrow$ M} is denoted as \textbf{B} and \textbf{S$\rightarrow$ M} is denoted as \textbf{C}.}
  \label{digits_table_semisup}
  \vspace{-1.0em}
  \end{center}
\end{table}

\textbf{Ablation Study: }We examined the performance of AVDA adopting two different training strategies, changing components of our framework that we consider critical. The idea is to verify how the performance of the model is affected if one of the components of the model is changed. We assumed that through domain adversarial learning we are able to capture domain-invariant features that allow us to improve the alignment in latent space. To prove this we tested the model removing the domain discriminator. We denoted this experiment as AVDA\textsubscript{WD}.
Second, we investigate how much does the pre-training process improve the performance of the model by testing AVDA without pre-training. We denote this experiment as AVDA\textsubscript{WP}. The experiments were performed in the most difficult digit scenario \textbf{S$\rightarrow$ M} using five labels per class. For the first experiment, AVDA\textsubscript{WD} obtained an accuracy of $91.57\pm1.02$, hence lowering the accuracy of our model.
For the second experiment, AVDA\textsubscript{WP} obtained an accuracy of $65.23\pm4.67$, notably decreasing the the model performance. In consequence, both domain adversarial process and pre-training makes AVDA obtain greater performance.


\begin{figure}[h]
\centering
  \includegraphics[width=1.0\linewidth]{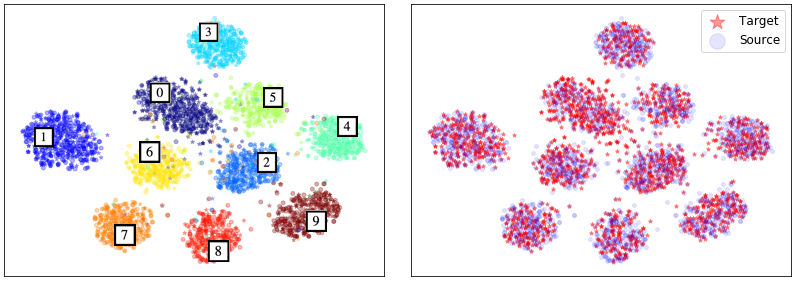}
  \caption{Visualization of the embedding space using t-SNE for the \textbf{M$\rightarrow$ U} task considering a 5-shot scenario. Colors on the left panel represent the data labels, showing the capability of the model to generate good discriminative boundaries. Colors on the right panel represent source and target data showing the capability of the model to align source and target domains into the same embedding components.}
  \label{fig:mu_tsne}
  \end{figure}

\begin{figure}[h]
\centering
  \includegraphics[width=1.0\linewidth]{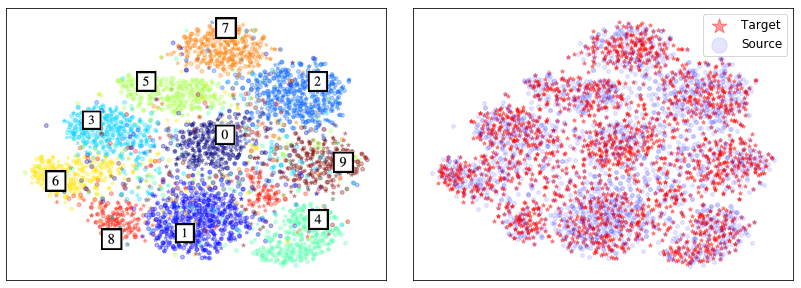}
\caption{Visualization of the embedding space using t-SNE for the \textbf{S$\rightarrow$ M} task considering a 5-shot scenario. Colors on the left panel represent the data labels, showing the capability of the model to generate good discriminative boundaries. Colors on the right panel represent source and target data showing the capability of the model to align source and target domains into the same embedding components.}
\label{fig:sm_tsne}
\end{figure}

\textbf{Visualization: } In order to visualize the alignment between source and target domains, we visualize the embedding space by using t-distributed stochastic neighbor embedding (t-SNE, \cite{maaten_2008}) for the tasks \textbf{M$\rightarrow$ U} and \textbf{S$\rightarrow$ M} considering a 5-shot scenario. Figure \ref{fig:mu_tsne} shows the visualization for \textbf{M$\rightarrow$ U} and Figure \ref{fig:sm_tsne} the visualization for \textbf{S$\rightarrow$ M}. On the left, we show each class in a different color, demonstrating the classifying capability of the model. On the right, we show the source and target in different colors, demonstrating the ability of the model to generate good alignments between the data labels and the embedded components for both domains.

\section{Conclusion}

In this paper we present Adversarial Variational Domain Adaptation (AVDA), a semi-supervised approach for domain adaptations problem were a vast annotated source domain is available but few labels from a target domain exist. Unlike previous methods which align source and target domains into a single common feature space, we use a variational embedding deined by a Gaussian Mixture Model and align samples that belong to the same class into the same embedding component using domain adversarial methods. Our model works as a classifier and also as a generative model that helps the classification. Experiments on a digits dataset are used to validate the proposed approach. Through this experimentation we show that our method outperform previous state-of-the-art from 0.08\% to 1.88\% of accuracy and also is capable to obtain competitive results even using fewer number of labeled samples on target.

\clearpage
%
%
\bibliographystyle{splncs04}
\bibliography{egbib}

\newpage
\section{Appendix}
In the following sections we provide the derivations that allow to compute $\mathcal{L}_{sup}$ and $\mathcal{L}_{unsup}$ from Equation \ref{eq:lsup_1} and Equation \ref{eq:lunsup_1} respectively. For this purpuse, Jiang et al. 2017 \cite{jiang_2016}, Lemma 1, will be used: Given two multivariate Gaussian distributions defined as:
\begin{equation}
    q(\mathbf{z}) = \mathcal{N}(\mathbf{z}|\tilde{\mu}, \tilde{\sigma}^{2}\mathbf{I}) \,\,\,\,\,\,\,\,\,\,\,\,\,\,\,\,\,\ p(\mathbf{z}) = \mathcal{N}(\mathbf{z}|\mu, \sigma^{2}\mathbf{I}), 
\end{equation}
it is possible to derive:
\begin{equation} \label{eq:lemma}
    \int_{\mathbf{z}} q(\mathbf{z})\log{p(\mathbf{z})}d\mathbf{z} \begin{aligned}[t] &= \int_{\mathbf{z}} \mathcal{N}(\mathbf{z}|\tilde{\mu}, \tilde{\sigma}^{2}\mathbf{I}) \log{\mathcal{N}(\mathbf{z}|\mu, \sigma^{2}\mathbf{I})} d\mathbf{z} \\ &= -\frac{1}{2}\sum_{j=1}^{J}\log{(2\pi\sigma_{j}^{2})} + \frac{\tilde{{\sigma}}^{2}_{j}}{\sigma_{j}^{2}}+\frac{(\tilde{\mu}_{j}-\mu_{j})^{2}}{\sigma^{2}_{j}},
    \end{aligned}
\end{equation}
where $\mu_{j}$, $\tilde{\mu}_{j}$, $\sigma^{2}_{j}$, $\tilde{\sigma}^{2}_{j}$ denotes the $j^{th}$ element o $\mu$, $\tilde{\mu}$, $\sigma^{2}$, $\tilde{\sigma}^{2}$ respectively. $J$ is the dimensionality of $\mathbf{z}$.

\subsection{Appendix A}
In this section we provide the derivation to compute $\mathcal{L}_{sup}$ in Equation \ref{eq:lsup_1}. This term is composed by the supervised source and target objectives as follows:
\begin{equation}
    \mathcal{L}_{sup} \begin{aligned}[t] & = \mathcal{L}^{s}_{sup} + \mathcal{L}^{t}_{sup} \\
    & = \begin{aligned}[t] &- \mathcal{L}^{s}_{ELBO}  + \alpha^{t}\mathbb{E}_{(\mathbf{x},y)\in \mathcal{D}^{s}}[-\log{q_{\phi}(y|\mathbf{x})}] \\ &- \mathcal{L}^{t}_{ELBO}  + \alpha^{t}\mathbb{E}_{(\mathbf{x},y)\in \mathcal{D}^{t}}[-\log{q_{\phi}(y|\mathbf{x})}]
    \end{aligned}
    \end{aligned}
\end{equation}
As we can see in Equation \ref{eq:Lelbo_s} and Equation \ref{eq:Lelbo_t}, $\mathcal{L}^{s}_{ELBO}$ and $\mathcal{L}^{t}_{ELBO}$ are similar objective functions, but differs only in the parameters $\theta$ and $\rho$ for the reconstruction term of the source and target data described by Equation 5 and Equation 8 respectively. Assuming a single data point which from either source or target, it is possible to generalize $\mathcal{L}_{sup}$ as follows:
\begin{equation} \label{eq:Lsup_gen}
    \mathcal{L}_{sup} \begin{aligned}[t] &= - \mathcal{L}_{ELBO} - \alpha \E_{(\mathbf{x},y) \in \mathcal{D}^{s}\cup\mathcal{D}^{t}}[\log q_{\phi}(y|\mathbf{x})] \\
    & = \begin{aligned}[t] & -\mathbb{E}_{q_{\phi}(\mathbf{z}|\mathbf{x})}[p(\mathbf{x}|\mathbf{z})] + \E_{q_{\phi}(\mathbf{z}|\mathbf{x})}[q(\mathbf{z}|\mathbf{x})]  \\ &- \E_{q_{\phi}(\mathbf{z}|\mathbf{x})}[p(\mathbf{z}|y)] - \alpha \E_{(\mathbf{x},y)}[\log q_{\phi}(y|\mathbf{x})]
    \end{aligned}
    \end{aligned}
\end{equation}
Each term of Equation \ref{eq:Lsup_gen} can be computed as follows:

\begin{description}
\item $\bullet$ $\E_{q(\mathbf{z}|\mathbf{x})}[p(\mathbf{x}|\mathbf{z})]$ : Assuming that $\textbf{x}$ is modeled by a multivariate Gaussian Distribution, we can derive the maximum likelihood estimation for the reconstruction term as follows:
\begin{equation} \label{eq:rec_sup}
     \E_{q_{\phi}(\mathbf{z}|\mathbf{x})}[p(\mathbf{x}|\mathbf{z})] \begin{aligned}[t] & =  \sum^{D}_{i=1} -\log{(2\pi\sigma^{2}_{x}|_{i})} - \frac{(x_{i} - \mu_{x}|_{i})^{2}}{2\sigma^{2}|_{i}},
    \end{aligned}
\end{equation}
where $\mu_{x}$ and $\sigma_{x}$ are the parameters for the Gaussian distribution of $\mathbf{x}$ and are computed using a neural network $f$ parametrized by $\theta$ if a sample comes from the source (see Equation 21) or $\rho$ if a sample comes from the target (see Equation 22). $\mathbf{z} \sim \mathcal{N}(\mu_{\phi}(\mathbf{x}), \sigma^{2}_{\phi}(\mathbf{x})\mathbf{I})$ and can be obtained by using the reparametrization trick defined by Equation \ref{reparam}. $D$ is the dimensionality of $\mathbf{x}$. \\

\item $\bullet$ $\E_{q_{\phi}(\mathbf{z}|\mathbf{x})}[q(\mathbf{z}|\mathbf{x})]$: By using the Lemma defined by Equation \ref{eq:lemma} and Equation 9, it is possible to rewrite this term as follows:
\begin{equation}
     \E_{q_{\phi}(\mathbf{z}|\mathbf{x})}[q(\mathbf{z}|\mathbf{x})] \begin{aligned}[t] & = \int_{\mathbf{z}} q_{\phi}(\mathbf{z}|\mathbf{x}) \log{q_{\phi}(\mathbf{z}|\mathbf{x})}d\mathbf{z} \\
     & = \int_\mathcal{N}(\mathbf{z}|\mu_{\phi}(\mathbf{x}), \sigma^{2}_{\phi}(\mathbf{x})\mathbf{I})\log{\mathcal{N}(\mathbf{z}|\mu_{\phi}(\mathbf{x}), \sigma^{2}_{\phi}(\mathbf{x})\mathbf{I}}) d\mathbf{z}
    \\ 
    & = -\frac{J}{2}\log{(2\pi)} -\frac{1}{2}\sum_{j=1}^{J}\big(1 + \log{\sigma^{2}_{\phi}(\mathbf{x})}|_{j}\big),
    \end{aligned}
\end{equation}
where J is the dimensionality of $\mathbf{z}$

\item $\bullet$ $\E_{q_{\phi}(\mathbf{z}|\mathbf{x})}[p(\mathbf{z}|y)]$: By using the Lemma defined by Equation \ref{eq:lemma} and Equations 9, 4 and 7, it is possible to rewrite this term as follows:

\begin{equation}
   \E_{q_{\phi}(\mathbf{z}|\mathbf{x})}[p(\mathbf{z}|y)] \begin{aligned}[t] &=  \int_{\mathbf{z}} q_{\phi}(\mathbf{z}|\mathbf{x}) \log p(\mathbf{z}|y)d\mathbf{z} \\ 
   & = \int_\mathcal{N}(\mathbf{z}|\mu_{\phi}(\mathbf{x}), \sigma^{2}_{\phi}(\mathbf{x})\mathbf{I})\log{\mathcal{N}(\mathbf{z}|\mu(y), \sigma^{2}(y})\mathbf{I}) d\mathbf{z}
   \\
   &= \begin{aligned}[t] & - \frac{J}{2}\log{(2\pi)} - \frac{1}{2} \sum_{j=1}^{J}\log{\sigma^{2}(y)}|_{j} 
   - \frac{1}{2} \sum_{j=1}^{J}\frac{\sigma^{2}_{\phi}(\mathbf{x})|_{j} }{\sigma^{2}(y)|_{j} } \\ & - \frac{1}{2} \sum_{j=1}^{J}\frac{(\mu_{\phi}(\mathbf{x})|_{j} -\mu(y)|_{j} )^{2}}{\sigma^{2}(y)|_{j}},
   \end{aligned}
  \end{aligned}  
\end{equation}
\\

\item $\bullet$ $\E_{(\mathbf{x},y)}[\log q_{\phi}(y|\mathbf{x})]$: The term that regularize the discriminative process of the model can be derived by using Equations 3 and 6 as follows:

\begin{equation}
    \E_{(\mathbf{x},y)}[\log q_{\phi}(y|\mathbf{x})] \begin{aligned}[t]  & = \sum_{k=1}^{K} y_{k}\log{q_{\phi}(y_{k}|\mathbf{x})} \\
    & = H(q_{\phi}(y|\mathbf{x}), y),
    \end{aligned}
\end{equation}
where $H$ is used to denote the categorical cross entropy loss. In practice, $q_{\phi}(y|\mathbf{x})$ is computed using $p(y|\mathbf{z})$ as is described in Equation 17. \\

Finally, it is possible to rewrite $\mathcal{L}_{sup}$ as follows:
\begin{equation}
    \mathcal{L}^{sup}  = 
    \begin{aligned}[t]  & \sum^{D}_{i=1} \log{(\sigma^{2}_{x}|_{i})} + \frac{(x_{i} - \mu_{x}|_{i})^{2}}{2\sigma^{2}|_{i}} \\
    & + \frac{1}{2} \Big(\sum_{j=1}^{J}\log{\sigma^{2}(y)}|_{j} 
   + \sum_{j=1}^{J}\frac{\sigma^{2}_{\phi}(\mathbf{x})|_{j} }{\sigma^{2}(y)|_{j} } + \sum_{j=1}^{J}\frac{(\mu_{\phi}(\mathbf{x})|_{j} + \mu(y)|_{j} )^{2}}{\sigma^{2}(y)|_{j}}\Big) \\
    & - \frac{1}{2}\sum_{j=1}^{J}\big(1 + \log{\sigma^{2}_{\phi}(\mathbf{x})}|_{j}\big) - \alpha H(q_{\phi}(y|\mathbf{x}),y)
    \end{aligned}
\end{equation}

\end{description}

\subsection{Appendix B}
Assuming a single unlabeled target data point, it is possible to derive $\mathcal{L}_{unsup}$ as follows:
\begin{equation} \label{eq:Lunsup_elbo}
  \mathcal{L}_{unsup} \begin{aligned}[t]  &= - \mathcal{L}^{u}_{ELBO} \\
 & = - \mathbb{E}_{q_{\phi}(y,\mathbf{z}|\mathbf{x})}[\log p_{\rho}(\mathbf{x}|\mathbf{z})] + D_{KL}(q_{\phi}\big(y,\mathbf{z}|\mathbf{x})||p(\mathbf{z},y)\big) \\
  &= \begin{aligned}[t] & - \mathbb{E}_{q_{\phi}(y,\mathbf{z}|\mathbf{x})}[\log p_{\rho}(\mathbf{x}|\mathbf{z})] + \mathbb{E}_{q_{\phi}(y,\mathbf{z}|\mathbf{x})}[\log q_{\phi}(\mathbf{z}|\mathbf{x})] \\ 
  &- \mathbb{E}_{q_{\phi}(y,\mathbf{z}|\mathbf{x})}[p(\mathbf{z}|y)]  +  \mathbb{E}_{q_{\phi}(y,\mathbf{z}|\mathbf{x})}[\log q_{\phi}(y|\mathbf{x})] - \mathbb{E}_{q_{\phi}(y,\mathbf{z}|\mathbf{x})}[\log p(y)]
  \end{aligned}
\end{aligned}
\end{equation}

Each term of Equation \ref{eq:Lunsup_elbo} can be computed as follows:

\begin{description}
\item $\bullet$ $\E_{q_{\phi}(\mathbf{z}|\mathbf{x})}[p_{\rho}(\mathbf{x}|\mathbf{z})]$ : Assuming that $\textbf{x}$ is modeled by a multivariate Gaussian Distribution, we can derive the maximum likelihood estimation for the reconstruction term similar as Equation \ref{eq:rec_sup} as follows:
\begin{equation}
     \E_{q_{\phi}(\mathbf{z}|\mathbf{x})}[p(\mathbf{x}|\mathbf{z})] \begin{aligned}[t] & =  \sum^{D}_{i=1} -\log{(2\pi\sigma^{2}_{x}|_{i})} - \frac{(x_{i} - \mu_{x}|_{i})^{2}}{2\sigma^{2}|_{i}},
    \end{aligned}
\end{equation}
where $[\mu_{x}, \sigma_{x}] = f(\mathbf{x}; \rho)$ are the parameters for the Gaussian distribution of $\mathbf{x}$ and are computed using a neural network $f$ parametrized by $\rho$.
$\mathbf{z} \sim \mathcal{N}(\mu_{\phi}(\mathbf{x}), \sigma^{2}_{\phi}(\mathbf{x})\mathbf{I})$ can be obtained by using the reparametrization trick defined by Equation \ref{reparam}. $D$ is the dimensionality of $\mathbf{x}$. \\

\item $\bullet$ $\mathbb{E}_{q_{\phi}(y,\mathbf{z}|\mathbf{x})}[\log q_{\phi}(\mathbf{z}|\mathbf{x})]$: By using the Lemma defined by Equation \ref{eq:lemma} and Equation 9, it is possible to rewrite this term as follows:

\begin{equation}
    \mathbb{E}_{q_{\phi}(y,\mathbf{z}|\mathbf{x})}[\log q_{\phi}(\mathbf{z}|\mathbf{x})] \begin{aligned}[t] & = \int_{\mathbf{z}} \sum^{K} q_{\phi}(y|\mathbf{x})q_{\phi}(\mathbf{z}|\mathbf{x}) \log{q_{\phi}(\mathbf{z}|\mathbf{x})}d\mathbf{z} \\
    & = \int_{\mathbf{z}}q_{\phi}(\mathbf{z}|\mathbf{x}) \log{q_{\phi}(\mathbf{z}|\mathbf{x})}d\mathbf{z} \\
    & = \int_\mathbf{z}(\mathbf{z}|\mu_{\phi}(\mathbf{x}), \sigma^{2}_{\phi}(\mathbf{x})\mathbf{I})\log{\mathcal{N}(\mathbf{z}|\mu_{\phi}(\mathbf{x}), \sigma^{2}_{\phi}(\mathbf{x})\mathbf{I}}) d\mathbf{z}
    \\ 
    & = -\frac{J}{2}\log{(2\pi)} -\frac{1}{2}\sum_{j=1}^{J}\big(1 + \log{\sigma^{2}_{\phi}(\mathbf{x})}|_{j}\big),
    \end{aligned}
\end{equation}

\item $\bullet$ $\mathbb{E}_{q_{\phi}(y,\mathbf{z}|\mathbf{x})}[p(\mathbf{z}|y)]$: By using the Lemma defined by Equation \ref{eq:lemma} and Equations 7 and 9, it is possible to rewrite this term as follows:
\begin{equation}
    \mathbb{E}_{q_{\phi}(y,\mathbf{z}|\mathbf{x})}[p(\mathbf{z}|y)] \begin{aligned}[t] & = \int_{\mathbf{z}} \sum^{K}_{k=1} q_{\phi}(y_{k}|\mathbf{x})q_{\phi}(\mathbf{z}|\mathbf{x}) \log{p(\mathbf{z}|y_{k})}d\mathbf{z} \\
   & = \sum^{K}_{k=1} q_{\phi}(y_{k}|\mathbf{x})  \int_\mathbf{z}(\mathbf{z}|\mu_{\phi}(\mathbf{x}), \sigma^{2}_{\phi}(\mathbf{x})\mathbf{I})\log{\mathcal{N}(\mathbf{z}|\mu_{\phi}(y), \sigma^{2}(y)}\mathbf{I}) d\mathbf{z} \\
   & = \begin{aligned}[t] &- \frac{1}{2}\sum^{K}_{k=1}q_{\phi}(y_{k}|\mathbf{x})\Big(\sum^{J}_{j=1} \log{(2\pi\sigma^{2}(y_{k})|_{j}))} + \sum^{J}_{j=1}\frac{\sigma^{2}_{\phi}(\mathbf{x})|_{j}}{\sigma^{2}(y_{k})|_{j}} \\ &+ 
   \sum^{J}_{j=1}\frac{(\mu_{\phi}(\mathbf{x})|_{j}-\mu(y_{k})|_{j})^{2}}{\sigma^{2}(y_{k})|_{j}} \Big),
   \end{aligned}
    \end{aligned}
\end{equation}
where $k \in {1,...,K}$ are the classes associated to each label $y$.

\item $\bullet$ $\mathbb{E}_{q_{\phi}(y,\mathbf{z}|\mathbf{x})}[\log q_{\phi}(y|\mathbf{x})]$: By using Equation 10, it is possible to derive this term as follows:

\begin{equation}
    \mathbb{E}_{q_{\phi}(y,\mathbf{z}|\mathbf{x})}[\log q_{\phi}(y|\mathbf{x})] \begin{aligned}[t] &= \int_{\mathbf{z}}\sum^{K}_{k=1} q_{\phi}(\mathbf{z}|\mathbf{x})q_{\phi}(y_{k}|\mathbf{x}) \log q_{\phi}(y_{k}|\mathbf{x}) d\mathbf{z} \\
    & = \sum_{k=1}^{K} q_{\phi}(y_{k}|\mathbf{x}) \log q_{\phi}(y_{k}|\mathbf{x})
    \end{aligned}
\end{equation}

\item $\bullet$ $\mathbb{E}_{q_{\phi}(y,\mathbf{z}|\mathbf{x})}[\log p(y)]$: By using Equations 6 and 10, it is possible to derive this term as follows:

\begin{equation}
   \mathbb{E}_{q_{\phi}(y,\mathbf{z}|\mathbf{x})}[\log p(y)] \begin{aligned}[t] &= \int_{\mathbf{z}}\sum^{K}_{k=1} q_{\phi}(\mathbf{z}|\mathbf{x})q_{\phi}(y_{k}|\mathbf{x}) \log p(y_{k}) d\mathbf{z} \\
    & = \sum_{k=1}^{K} q_{\phi}(y_{k}|\mathbf{x}) \log p(y_{k})
    \end{aligned}
\end{equation}

Finally, $\mathcal{L}_{unsup}$ can be rewitten as:
\begin{equation}
\mathcal{L}_{unsup} \begin{aligned}[t] =  & \sum^{D}_{i=1} \log{\sigma_{x}|_{i}} + \frac{(x_{i} - \mu_{x}|_{i})^{2}}{2\sigma^{2}|_{i}} \\
&  + \frac{1}{2}\sum^{K}_{k=1}q_{\phi}(y_{k}|\mathbf{x})\sum^{J}_{j=1} \Big(\log{\sigma^{2}(y_{k})|_{j})} + \frac{\sigma^{2}_{\phi}(\mathbf{x})|_{j}}{\sigma^{2}(y_{k})|_{j}} + \frac{(\mu_{\phi}(\mathbf{x})|_{j}-\mu(y_{k})|_{j})^{2}}{\sigma^{2}(y_{k})|_{j}} \Big) \\
& -\sum^{K}_{k=1}q_{\phi}(y_{k}|\mathbf{x}) \log{\frac{\log p(y_{k})}{q_{\phi}(y_{k}|\mathbf{x})}} - \sum^{J}_{j=1}(1+\log \sigma^{2}_{\phi}(\mathbf{x})|_{j}).
\end{aligned}
\end{equation}

To calculate Equation 41, we used $q_{\phi}(y|\mathbf{x}) = p(y|\mathbf{z})$ as is described in Equation 17. \\

\end{description}

\end{document}